\documentclass[conference]{IEEEtran}

\usepackage[utf8]{inputenc} 
\usepackage[T1]{fontenc}    
\usepackage{hyperref}       
\usepackage{url}            
\usepackage{booktabs}       
\usepackage{amsfonts}       
\usepackage{nicefrac}       
\usepackage{microtype}      
\usepackage{lipsum}	     	
\usepackage{graphicx}
\usepackage{doi}
\usepackage{natbib}
\usepackage{amsmath}
\usepackage{todonotes}
\usepackage{arydshln}
\usepackage{caption}
\usepackage{subcaption}

\usepackage{soul}
\IEEEoverridecommandlockouts

\title{A Graph-based U-Net Model for Predicting Traffic in unseen Cities}

\author{ 
\href{https://orcid.org/0000-0002-7568-7981}{\includegraphics[scale=0.06]{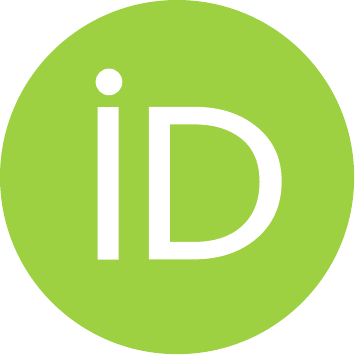}\hspace{1mm}Luca Hermes}$^1$, 
\href{https://orcid.org/0000-0002-0935-5591}{\includegraphics[scale=0.06]{orcid.pdf}\hspace{1mm}Barbara Hammer}$^1$, 
\href{https://orcid.org/0000-0002-7252-9267}{\includegraphics[scale=0.06]{orcid.pdf}\hspace{1mm}Andrew Melnik}$^2$, 
\href{https://orcid.org/0000-0002-2160-4976}{\includegraphics[scale=0.06]{orcid.pdf}\hspace{1mm}Riza Velioglu}$^1$, 
\href{https://orcid.org/0000-0003-1707-6231}{\includegraphics[scale=0.06]{orcid.pdf}\hspace{1mm}Markus Vieth}$^1$, 
\href{https://orcid.org/0000-0002-0849-483X}{\includegraphics[scale=0.06]{orcid.pdf}\hspace{1mm}Malte Schilling}$^1$\\
	\href{}{$^1$}Machine Learning Group, Bielefeld University \\
	\href{}{$^2$}Neuroinformatics Group, Bielefeld University \\
    33615 Bielefeld, Germany
	\thanks{This research was supported by the research training group ``Dataninja'' (Trustworthy AI for Seamless Problem Solving: Next Generation Intelligence Joins Robust Data Analysis) funded by the German federal state of North Rhine-Westphalia.} \\

}

\date{}

\hypersetup{
pdftitle={A Graph-based U-Net Model for Predicting Traffic in unseen Cities},
pdfsubject={Traffic4Cast Challenge 2021},
pdfauthor={Luca Hermes, Andrew Melnik, Riza Velioglu, Markus Vieth, Malte Schilling},
pdfkeywords={Traffic4Cast, Traffic Prediction, Traffic Forecasting, Graph U-Net, Hybrid U-Net},
}

\begin{document}
\maketitle
\begin{abstract}
Accurate traffic prediction is a key ingredient to enable traffic management like rerouting cars to reduce road congestion or regulating traffic via dynamic speed limits to maintain a steady flow. One way to represent traffic data is as temporally changing heatmaps visualizing attributes of traffic, such as speed and volume.
In recent approaches, U-Net models have shown state of the art performance on traffic forecasting from such heatmaps. We propose to combine the U-Net architecture with graph layers which improves spatial generalization to unseen road networks compared to a Vanilla U-Net. In particular, we specialize existing graph operations to be sensitive to geographical topology and generalize pooling and upsampling operations to be applicable to graphs.
\end{abstract}


\section{Introduction}
In this work, we present our results from participating in the NeurIPS \textit{Traffic4Cast} (T4C) Challenge 2021 \cite{t4c_2021}. The competition aims to predict the development of traffic volume and speed up to one hour into the future. This task is challenging due to the stochastic nature of moving cars and complex spatio-temporal dependencies.
The T4C challenge 2021 was subdivided into two challenges with a focus on temporal and spatial transfer (the exact settings are described in detail in Sec. \ref{sec:challenge_details}).

Our approach is inspired by the well-known U-Net architecture \cite{ronnebergerUNetConvolutionalNetworks2015}, as used frequently in the scope of the competition. Although U-Net models have shown to perform well, transfer to unseen cities has been difficult for such convolution-based approaches. \cite{pmlr-v123-martin20a} provided empirical evidence that graph-based models generalize better to unseen cities as these allow to leverage prior knowledge about the street network.
Thus, instead of relying on visual convolutions (CNN), we apply graph neural networks (GNN) to integrate local traffic information using a road graph. One common drawback of using GNN-based approaches is the limited control over the receptive field. Expansion of the receptive field onto larger areas of the graph requires deeper models, however, it has been shown by \cite{DeepGNNsPerformanceDecay} that performance of GNNs drops at a certain depth.
We therefore adapted the visual pooling and upsampling operation in a way that they can account for long-range spatial relations between different areas in the road graph. To enable reproducibility, the code will be made publicly available\footnote{\url{https://github.com/LucaHermes/graph-UNet-traffic-prediction}}.


\section{Challenge Details and Data}
\label{sec:challenge_details}

\begin{figure*}[h!]
    \centering
    \includegraphics[width=\textwidth]{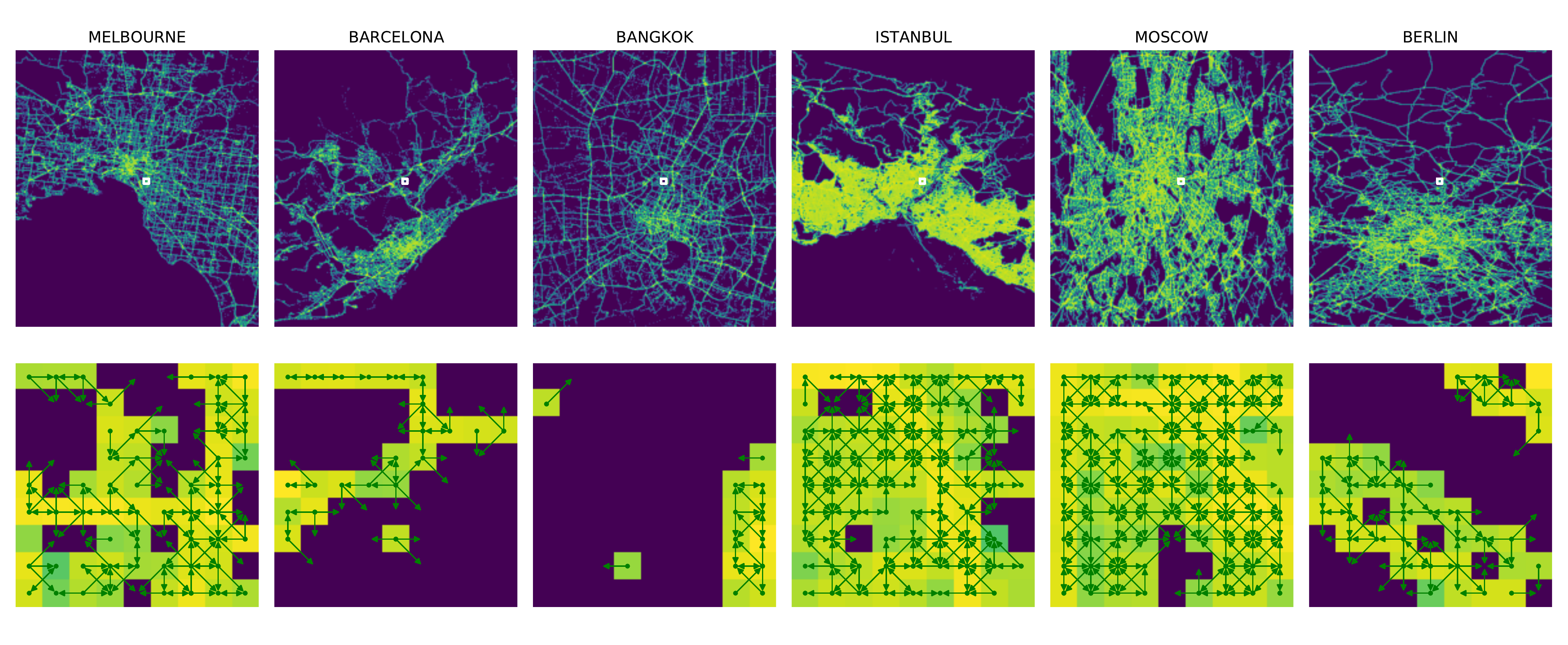}
    \caption{
    Sum over 24 consecutive frames (2 hours) of the traffic movies for six different cities on 24.04.2019 starting at 13:00. The color values are the sum over the speed and volume channels and are scaled logarithmically for better visibility. Top row are full size images ($495 \times 436$), white rectangles mark $8 \times 8$ windows. Bottom row shows these $8 \times 8$ windows with travel directions (green arrows).
    }
    \label{fig:demo_cities}
\end{figure*}

The traffic data is given as a two-dimensional heat-map-like image representation of size $495 \times 436$ pixels with eight channels.
The channels of the pixel values correspond to directional traffic speed and volume information binned into four discrete directions (north-east, south-east, south-west, and north-west). 
The data was sampled in a five-minute interval and was gathered from a fleet of probe vehicles. The measurements are mapped onto a pixel grid using GPS information to represent traffic movies as shown in Fig. \ref{fig:demo_cities}. 
Data was collected over two years (2019 and 2020) in ten different cities around the world. The traffic situation between 2019 and 2020 is subject to temporal shift caused by the ongoing COVID-19 pandemic which poses a particular challenge for transfer of models. In addition to the traffic movies, a road graph was generated from high-resolution images of the respective street network. The nodes in the graph correspond to pixels that belong to a street and an edge exists between two nodes if the corresponding pixels are adjacent and belong to a street. Furthermore, static street map images with a resolution similar to the traffic movies are also included for each city. These street maps are one-channel images, where the pixel intensities roughly correspond to street size.

\begin{table}[h!]
    \centering
    \caption{The different data subsets used in the core and extended (ext.) challenge.}
    \label{tab:data-splits}
    \resizebox{\linewidth}{!}{
    \begin{tabular}{l ll l}
    \hline
     Subset & 2019           & 2020                & Cities \\  \toprule
     C1     & train  & train       & Antwerp, Bangkok, Barcelona, Moscow  \\ 
     C2     & train  & -                   & Berlin, Chicaco, Istanbul, Melbourne \\
     C3     & -              & test (core) & Berlin, Chicaco, Istanbul, Melbourne \\
     C4     & test (ext.) & test (ext.) & New York, Vienna \\
    \hline
    \end{tabular}
    }
\end{table}

This year's challenge is divided into a core challenge and an extended challenge. The core challenge puts a focus on temporal generalization regarding the domain shift caused by COVID, while the extended challenge puts a focus on spatial generalization to unseen cities. Participants were invited to compete in both independently.
Four different subsets are derived from the data (s. Tab. \ref{tab:data-splits}) and define the two challenges. Specifically, both challenges are using the same training dataset (subsets C1 and C2), but different test sets. The evaluation of the core challenge focuses on generalization from pre-COVID training data of 2019, to the test set C3 recorded during COVID in 2020. The extended challenge focuses on generalization across cities. Model evaluation for this challenge uses the test set C4 that consists of data from two cities that were excluded from the training set. The task in both challenges is to predict the traffic 5, 10, 15, 30, 45, and 60 minutes into the future.

\section{Related Work}
\label{sec:related_work}
Traffic data can be generated by a variety of methods with different data representations. A common data acquisition method is to use stationary road sensors that measure attributes of passing traffic like speed and volume. The PEMS-BAY and METR-LA datasets \cite{big_data_metr_la} are popular examples. This data can be organized in a graph, where each sensor corresponds to a node and edges are inserted based on node proximity.
A second method is to collect data from mobile sensors, i.e. GPS-equipped cars. This allows to densely capture traffic attributes over large areas. The latter was used to collect the T4C dataset as mentioned in Sec. \ref{sec:challenge_details}.
The two methods produce data with different characteristics and, therefore, naturally lead to two broad groups of traffic prediction approaches that use data from either stationary, or mobile sensors.

In the first group that employs datacaptured from stationary sensors, spatial dependencies in the data are commonly captured with a graph neural network to process the irregular spatial dependencies of the road sensors. 
\cite{li2018_dcrnn} used a GRU in combination with a novel diffusion graph convolution. The latter is an extension to the vanilla graph convolution that allows to propagate information farther across the graph. \cite{graph_wavenet} proposed to combine the diffusion graph convolutions with dilated 1-dimensional convolutions arranged according to the WaveNet schematic \cite{wavenet}, thereby strictly limiting the temporal receptive field. \cite{Zhang_SLCNN_2020} learned the relations between nodes in the road network on global and local scales and apply graph convolutions to these learned graphs, effectively enhancing the latent node embeddings.

When considering information from mobile sensors, as done in the second group, the data is often represented as sequences of images. This is achieved by mapping GPS trajectories onto a discrete grid. Therefore, spatial dependencies can be modeled with image processing techniques but GNNs could also be applied. \cite{jo_img2img_2019} proposed a convolutional enconder-decoder model to predict the next frame of traffic.
The model receives multiple frames as image channels to capture temporal dependencies directly which outperforms a variant with an RNN. This way of temporal processing has been predominant in the contributions to past T4C challenges. The winning solution of 2019 \cite{t4c_choi_2019} organized blocks of densely connected convolutions \cite{gao_dense_conv} in a U-Net-style architecture \cite{ronnebergerUNetConvolutionalNetworks2015}. \cite{wu_2020_hrnet} showed that the HR-Net architecture \cite{t4c_2020_wang} preduced promising results as well.
In contrast, \cite{pmlr-v123-martin20a} proposed Graph-ResNet, a deep graph architecture with residual connections. To apply this graph-based model, they converted the sparse images into graphs and reported better spatial generalization to unseen cities compared to convolutional approaches.

This work is most related to the Graph-ResNet approach, however, we organize our graph operations in a U-Net architecture. Additionally, we focus on improving the spatial dependencies by not only using the graph topology, but also the geographical topology, as described in Sec. \ref{sec:traffic_prediction}.







\section{Traffic Prediction}
\label{sec:traffic_prediction}
\begin{figure}
    \centering
     \begin{subfigure}[b]{0.49\textwidth}
         \centering
         \includegraphics[width=\textwidth]{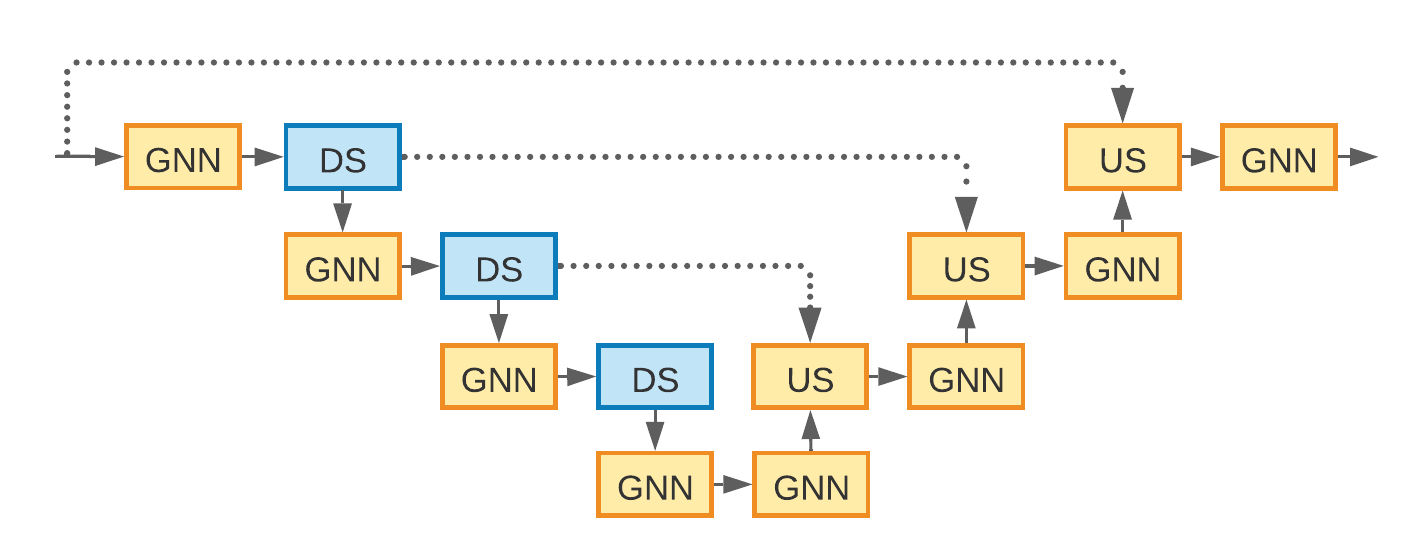}
         \caption{}
         \label{fig:model_architecture}
     \end{subfigure}
     \hfill
     \begin{subfigure}[b]{0.49\textwidth}
         \centering
         \includegraphics[width=\textwidth]{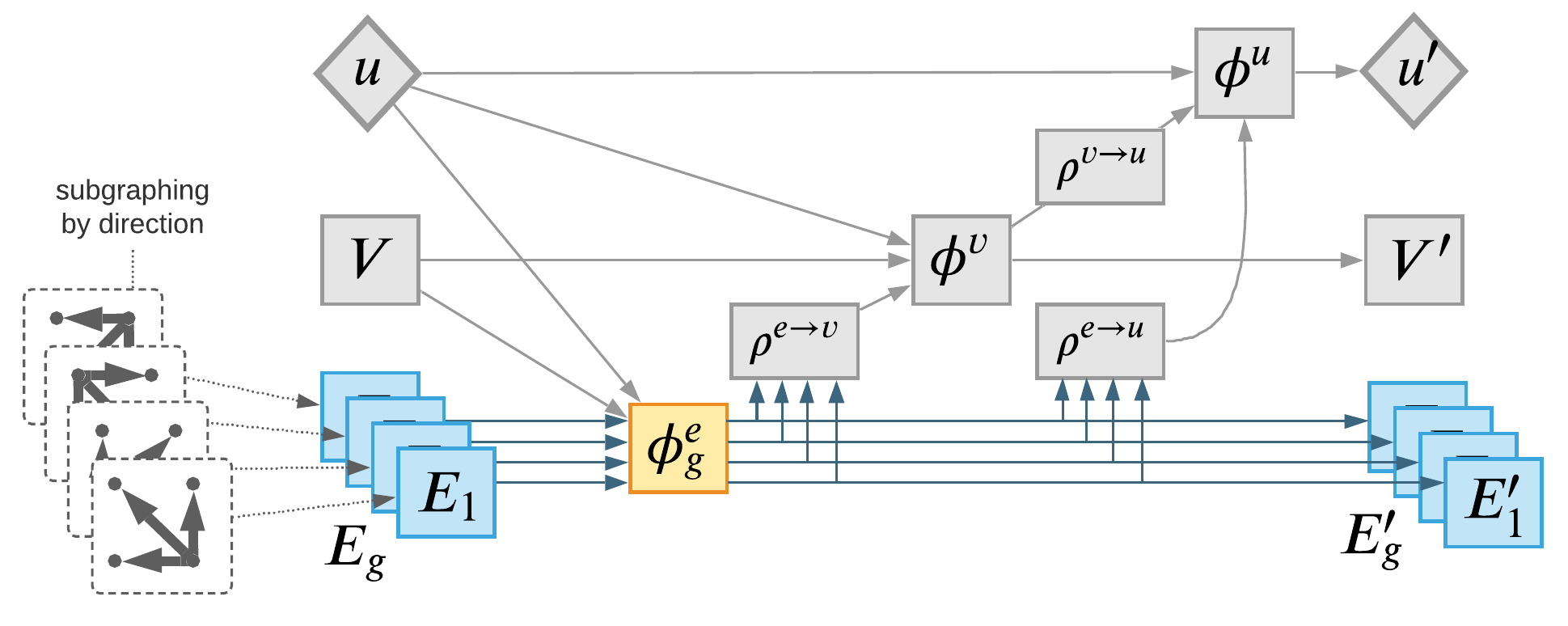}
         \caption{}
         \label{fig:fullGNBlock}
     \end{subfigure}
     \caption{(a) Overview of the graph-based U-Net architecture.
     Dotted lines represent skip connections that are concatenated to the channels at the end; DS denotes downsampling operations; US denotes upsampling operations. Right: Computations inside a GNN layer. Data flows from left to right, $E_g$ denotes edge subsets, $V$ denotes the set of vertices and $u$ denotes the global state vector. Fig. (b) is inspired by Fig. 4 (a) of \cite{battagliaRelationalInductiveBiases2018}.
     Our changes to the original \textit{full GN Block}
     are highlighted as colored elements.
     The operations $\phi$ are functions that can be learned. Elements in (a) that contain the GNN Layer (b) are denoted by yellow boxes.}
\end{figure}

Our architecture is based on the classical U-Net model \cite{ronnebergerUNetConvolutionalNetworks2015}, which originally relies on a series of two-dimensional visual convolutions. Thus, intermediate feature maps depend not only on the traffic data but also on city-specific empty areas in the traffic movies. 
As the road graph provides more specific topological information than just a regular pixel grid, we generalize this model to graphs by applying GNN layers instead of visual convolutions (CNN).
As a consequence, empty areas are excluded from computations. Thereby, these areas cannot affect the downstream latent representations, which we assume as beneficial for cross-city generalization.

In contrast to CNNs, GNNs cannot capture the geographical topology of a node neighborhood due to the permutation invariant accumulation functions. For example, a regular GNN cannot distinguish whether a neighboring node lies to the north or the south. Invariance to geographical topologies seems like a major drawback, as the traffic features are directional.
We propose a straightforward method to mitigate this drawback and specialize a GNN to be sensitive to these geographical neighborhood topologies (s. Sec. \ref{sec:graph_convolution}).
As the edges in the road graph only ever connect nodes that are also adjacent in the image space, the graph diameter (maximum distance between pixels on the graph) is $d \geq 495 + 436 = 931$.
A single-layer GNN can explore a 1-hop neighborhood, thus, long-range relations between nodes could only be exploited using a very deep GNN. However, it has been shown by \cite{DeepGNNsPerformanceDecay} that common GNNs don't scale well with model depth and tend to oversmooth in such cases.
To still allow information exchange over the whole graph, we include a global state vector in our GNN layers. This vector can be understood as being adjacent to every node in the graph. 
Furthermore, we leverage the unique position in the 2D pixel grid of each node, to design down- and upsampling operations (s. Sec. \ref{sec:down_up_sampling}), and thus to expand the receptive field of the GNN.
Following the U-Net schematic depicted in Fig. \ref{fig:model_architecture}, 
we arrange these operations in a down- and an upsampling branch that are additionally connected via skip-connections. The GNNs in the downsampling branch consist of a single layer, whereas the GNNs in the upsampling branch consist of two layers.
We will first describe the different types of features used in our layers and then the computations inside the layers.

\subsection{Feature Generation}
\label{sec:feature_engineering}
The input to our model is composed of the given road graph and the static street map images. The road graph is extended by edge feature vectors $\mathbf{e}_k \in E$ and one global feature vector $\mathbf{u}$.
The node features $\mathbf{v}_i \in \mathcal{V}$ correspond to the pixel values of the traffic movies, where the individual frames are concatenated into a single vector per node. Speed and volume information is scaled down to values between 0 and 1, i.e. divided by 255.
To initialize edge features, a two-layer CNN generates a feature map with eight channels from the normalized static street map $S \in \mathbb{R}^{1 \times h \times w \times 1}$. This results in an additional set of node features $\mathcal{\widetilde{V}}$ and by concatenating sender and receiver nodes from $\mathcal{\widetilde{V}}$ yield the edge features

\begin{align}
    \mathcal{\widetilde{V}} &= \text{CNN}(S)\\
    \mathbf{e}_k &= \left[\mathbf{\widetilde{v}}_{sk} \mathbin\Vert \mathbf{\widetilde{v}}_{rk} \right],
\end{align}

where $\mathbf{\widetilde{v}}_{sk}$ and $\mathbf{\widetilde{v}}_{rk}$ are the sender and receiver node of edge $k$, respectively, and $\cdot \mathbin\Vert \cdot$ denotes concatenation.
The global state is computed by summing up the node features $\mathcal{V}$ and scaling them by a constant $\lambda = 1 \times 10^{-5}$. This factor has to be included, as the sum over all nodes can be large and scaling by $\lambda$ showed empirically to produce suitable numbers. Next, time and weekday information is encoded in $\mathbf{t}_{enc}$ and $\mathbf{d}_{enc}$, respectively, and concatenated to the global state vector. Time $t$ is encoded as a 2D position on the unit circle, where the 24-hour interval corresponds to one full revolution and weekday is one-hot encoded.


\begin{align}
    \mathbf{u} &= \left[  \mathbf{\hat{v}}  \mathbin\Vert  \mathbf{t_{enc}}(t)  \mathbin\Vert  \mathbf{d_{enc}}(d)  \right] \\
    \mathbf{\hat{v}} &= \lambda \sum_{\forall \mathbf{v_i} \in \mathcal{V}} \mathbf{v_i} \\
    \mathbf{t_{enc}}(t) &= \left[  \sin{t}  \mathbin\Vert  \cos{t}  \right]
\end{align}

\subsection{Graph Layer}
\label{sec:graph_convolution}

\begin{figure*}
    \centering
    \includegraphics[width=.8\linewidth]{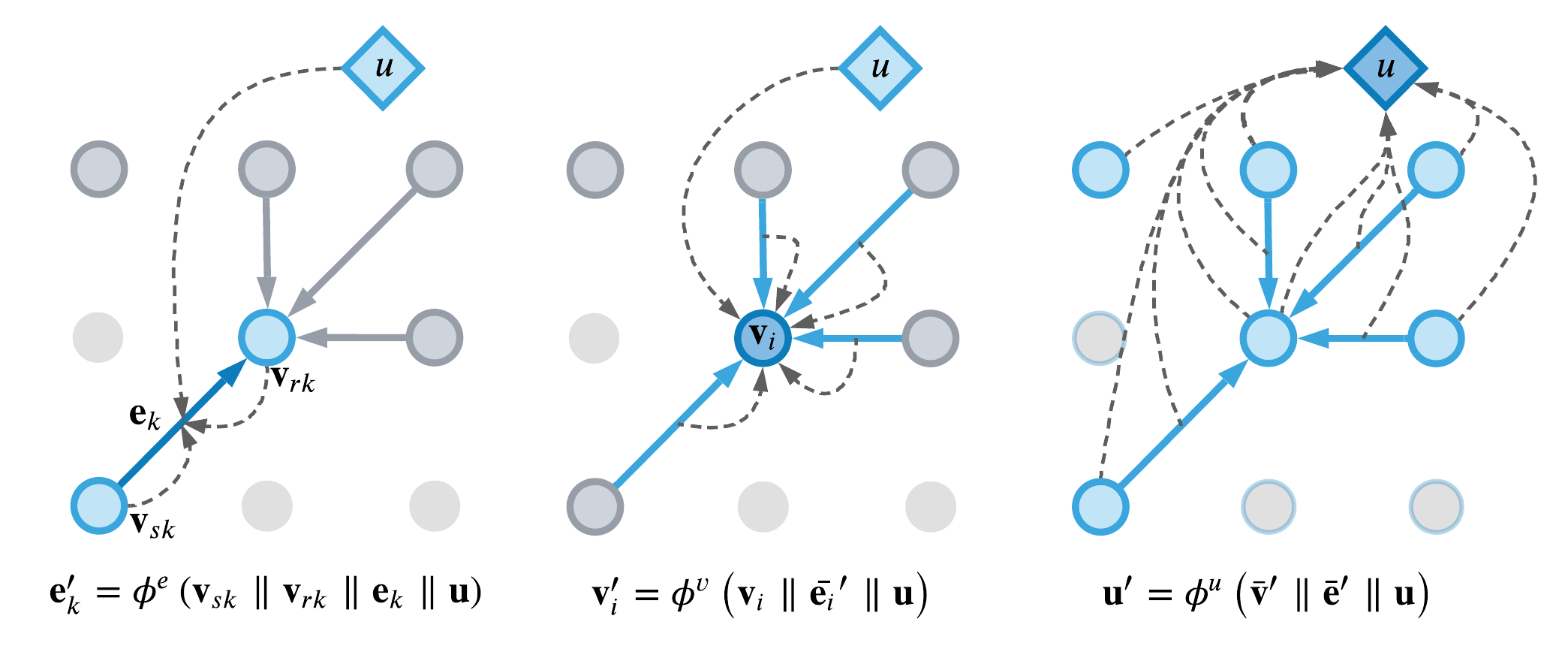}
    \caption{Edge update (left), node update (middle) and global update (right) of the \textit{full GN Block} \cite{battagliaRelationalInductiveBiases2018}. The inputs to the respective function are highlighted as the light blue elements, outputs are highlighted as dark blue elements, elements that are not involved in the computation are gray. $\cdot \parallel \cdot$ denotes concatenation.}
    \label{fig:fullGNBlock_visualization}
\end{figure*}


As basis of our graph layer, we use the \textit{full GN block} proposed by \cite{battagliaRelationalInductiveBiases2018}, as shown in Fig. \ref{fig:fullGNBlock}. It is composed of parameterized update functions $\phi$ that are applied to update node, edge, and global graph features, as well as unparameterized functions $\rho$ that accumulate sets. These functions are implemented as

\begin{align}
\phi(\mathbf{a}_1,, ..., \mathbf{a}_n) &= \text{ReLU}\left( \mathbin\Vert_{\forall \mathbf{a}_i} \mathbf{a}_i \, \mathbf{W} + \mathbf{b}\right) \\
\rho(A) &= \sum_{\mathbf{a}_i \in A} \mathbf{a}_i,
\end{align}

where $\mathbin\Vert_{\forall \mathbf{a}_i}$ denotes vector concatenation over all input vectors $\mathbf{a}_i$, $\mathbf{W}$ is the weight matrix and $\mathbf{b}$ is the bias vector. \textit{ReLU} \cite{Agarap2018DeepLU} is used as the activation function. Fig. \ref{fig:fullGNBlock_visualization} visualizes the computations conceptually.





\begin{figure*}
    \centering
    \includegraphics[width=\linewidth]{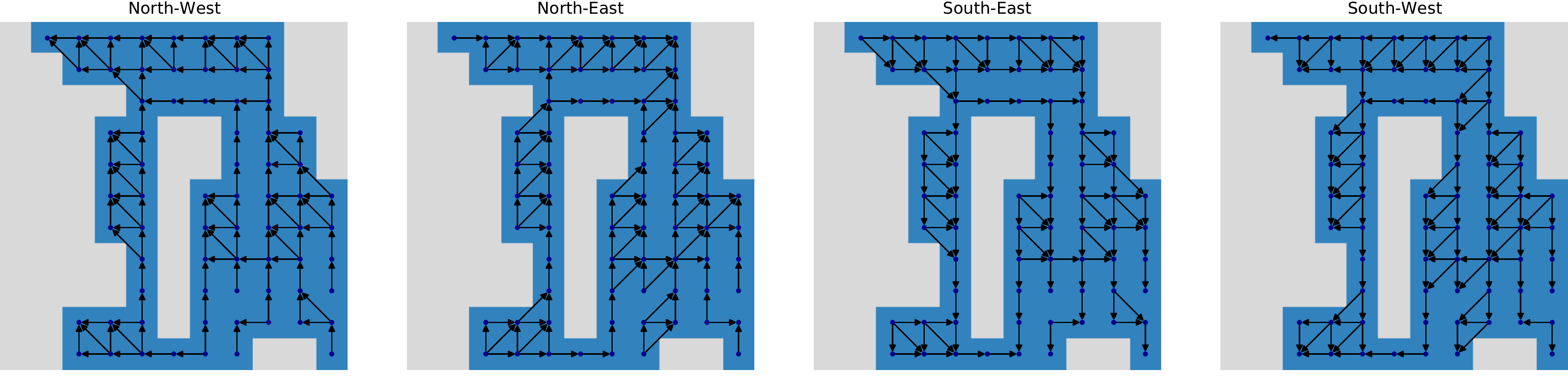}
    \caption{The four directed subgraphs that we extracted from the road graph. The graphs only contain edges in the respective geographical direction. The plots show a fragment of the road graph of Berlin. Blue pixels denote a street, gray pixels denote no street.}
    \label{fig:subgraphs}
\end{figure*}

We adapt the \textit{full GN block} to make the computations sensitive to the local neighborhood topology (adaptations highlighted in Fig. \ref{fig:fullGNBlock}). Specifically, we split the road graph into four subgraphs $g \in G$. As shown in Fig. \ref{fig:subgraphs} the subgraphs each contain edges directed into one of the four quadrants \textit{north-west}, \textit{north-east}, \textit{south-east} and \textit{south-west}. We choose these four subgraphs, as they reflect the partitioning of the directional traffic speed and volume information as given in the node features. Note that each subgraph uses the same node features. For each subgraph $g$, separate edge transformations $\phi^e_g$ compute the updated edge features $\mathbf{e}_{g,k}'$ for each edge $k$ in graph $g$. Then, the updated node features $\mathbf{v}_{i}'$ are computed by concatenating the edge features of the four subgraphs and for each node $\mathbf{v}_i \in V$ accumulating the incident edges. Finally, the global state vector is updated using the accumulated node and edge features, as well as the prior global state vector as inputs. The update functions are

\begin{align}
\begin{split}
    \mathbf{e}_{g,k}'
     &= \phi^e_g(\mathbf{e}_{g,k}, \mathbf{v}_{rk}, \mathbf{v}_{sk}, \mathbf{u})  \\
     \mathbf{e}_{k}'
     & = \mathbin\Vert_{\forall g \in G} \, \mathbf{e}_{g,k} \\
     \mathbf{v}_{i}'
     &= \phi^v\left(\mathbf{v}_{i}, \mathbf{\bar{e}}_{i}', \mathbf{u}\right)\\
     \mathbf{u}'
     &= \phi^u\left(\mathbf{u}, \mathbf{\bar{v}}', \mathbf{\bar{e}}'\right)
\end{split}
\begin{split}
     \\
    \mathbf{\bar{e}}_{i}'
     &= \rho^{e \rightarrow v}(\{ \mathbf{e}_{k}' |  \mathbf{e}_{k}' \in \mathcal{N}_i\})  \\
     \mathbf{\bar{e}}'
     &= \rho^{e \rightarrow u}(E') \\
     \mathbf{\bar{v}}'
     &= \rho^{v \rightarrow u}(\mathcal{V}')
\end{split}
\end{align}

where $\mathbf{v}_{sk}$ and $\mathbf{v}_{rk}$ are the sender and receiver node of edge $k$, respectively, and $\mathcal{N}_i$ denotes the neighborhood of node $i$. The outputs of the graph layer are the new state vectors for nodes $\mathbf{v}_i'$, edges $\mathbf{e}_k'$ and the global features $\mathbf{u}'$.



\subsection{Downsampling and Upsampling}
\label{sec:down_up_sampling}
\begin{figure}
    \centering
    \includegraphics[width=\linewidth]{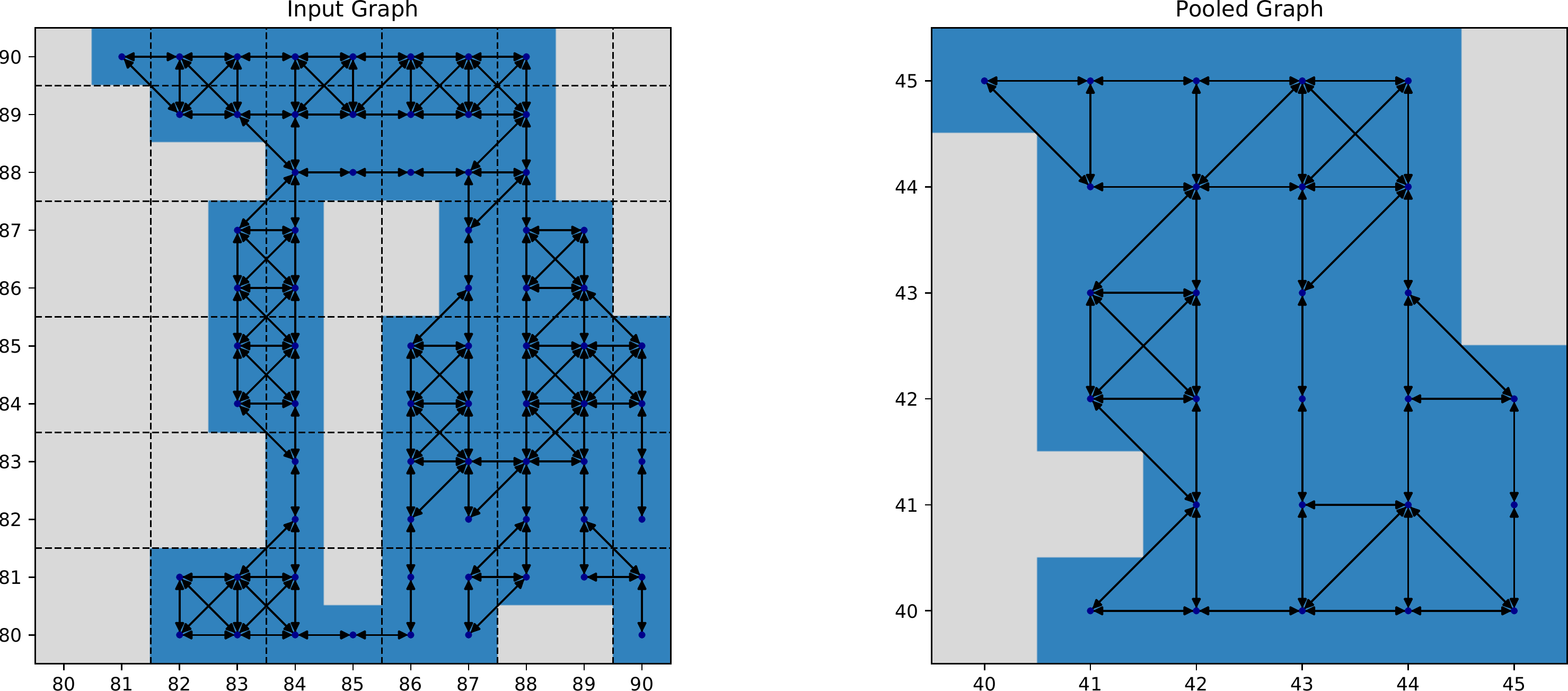}
    \caption{Pooling operation input (left) and output graph (right). Dashed lines show the pooling windows. Black axis ticks denote pixel position, blue axis ticks denote pooling window position.}
    \label{fig:graph_pooling}
\end{figure}

While down- and upsampling operations on graphs are not straightforward,  in the current setting we can exploit the topological information that locates nodes in the pixel grid. Thus, existing visual pooling and upsampling methods can be adapted to work on the street graph.
Our downsampling operation directly corresponds to a regular max-pooling with a kernel size and a stride of $2$. Specifically, we partition the set of nodes $\mathcal{V}$ according to their position in the 2D grid in a way that each partition contains the nodes in a $2 \times 2$ window and take the feature-wise maximum. Hence, each partition is condensed into a new node, resulting in a new set of nodes $\mathcal{V}'$. An edge connects two nodes $u'$ and $v'$ if any two nodes in the corresponding pooling windows were connected by an edge. If multiple such connections are present, the feature-wise maximum is taken to yield the new edge features. Finally, the new node position corresponds to the 2D index of the corresponding pooling window.
The edge features are updated by taking the feature-wise maximum over all edges that are mapped onto the same edge $e' \in E'$.

\begin{figure}
    \centering
    \includegraphics[width=\linewidth]{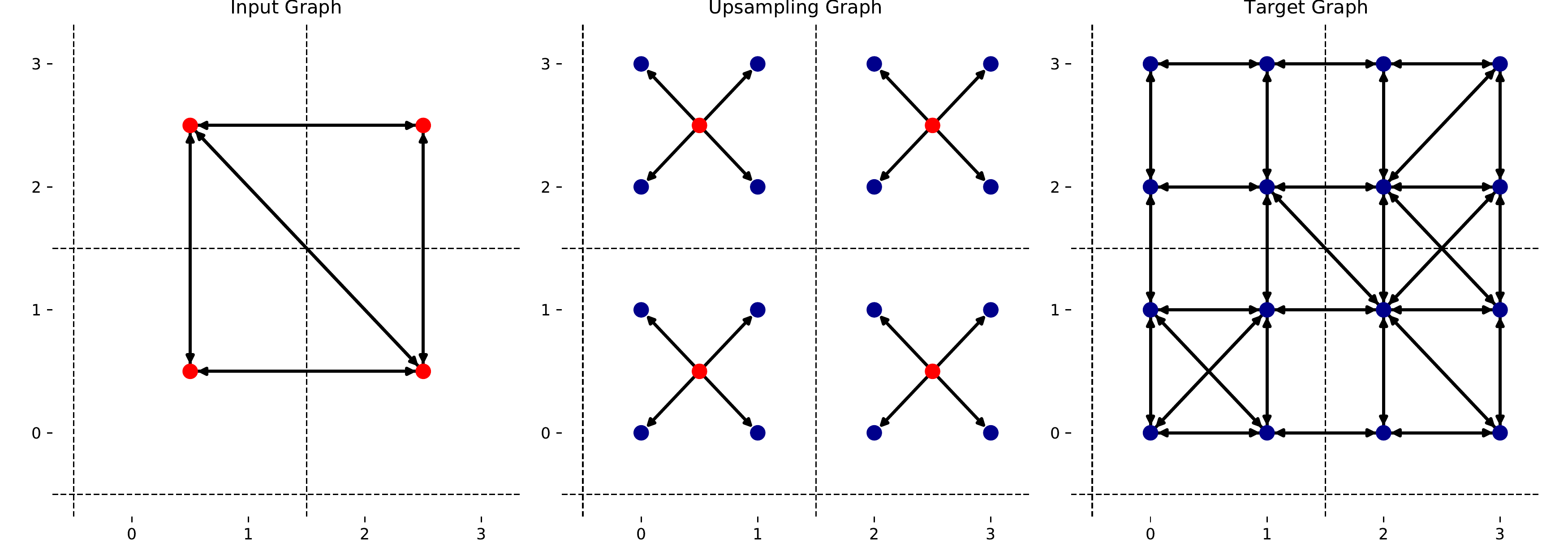}
    \caption{Upsampling strategy used to transform an input graph $G_i$ (left; red nodes) into a target graph $G_t$ (right; blue nodes). The node features in the upsampled graph are generated via a graph layer using the upsampling graph (middle). For better visibility, the input graph has been moved by $0.5$ in x- and y-direction.}
    \label{fig:graph_upsampling}
\end{figure}

The upsampling operation relies on a given input graph $G_i$ and a target graph structure $G_t$. Fig. \ref{fig:graph_upsampling} visualizes our upsampling method. First, $n_t$ zero-initialized nodes are introduced to the input graph, where $n_t$ denotes the number of nodes in $G_t$. Second, the position of input nodes is scaled by a factor of two. Then, we partition the nodes by their position in the two-dimensional space in a way that each partition contains the nodes in a $2 \times 2$ window, like in the downsampling case. Next, edges are created by connecting nodes in $G_i$ to all nodes in $G_t$ that are in the same partition. This creates an upsampling graph as shown in Fig. \ref{fig:graph_upsampling} (middle), effectively connecting $G_i$ with $G_t$. Our adapted GNN is applied to the upsampling graph to propagate information from $G_i$ to $G_t$. 
This results in new node features for the upsampled graph. Note that the GNN uses the same edge partitioning as described in Sec. \ref{sec:graph_convolution}. Hence, the operation is sensitive to the relative neighborhood topology and will therefore produce different values in the receiving nodes, even if the sending node is the same. 

\subsection{Training Setup}
We train our model for 800k steps on the provided training data on a standard MSE loss using the ADAM optimizer \cite{kingmaAdamMethodStochastic2017}. The learning rate schedule contains 2k steps of warm-up. After warm-up, the learning rate is  $0.002$ and decays from there exponentially at a rate of $0.98$ every 100 steps. The minimal learning rate is $0.0002$. At each step, we sample valid starting frames for the seed sequences (model input), which are frames that correspond to a time between 00:00 and 22:00 o'clock. We take the average over the gradient of 16 successive samples to update the model parameters. This effectively corresponds to taking a batch size of 16. We submitted the exact same model to both competitions. Hence, we consider the temporal generalization problem from the core challenge as a kind of spatial generalization problem as well. This is possible as data from 2020 is included in the training set, just for different cities than used in the evaluation.

\section{Results}
\begin{figure*}
    \centering
    \includegraphics[width=\linewidth]{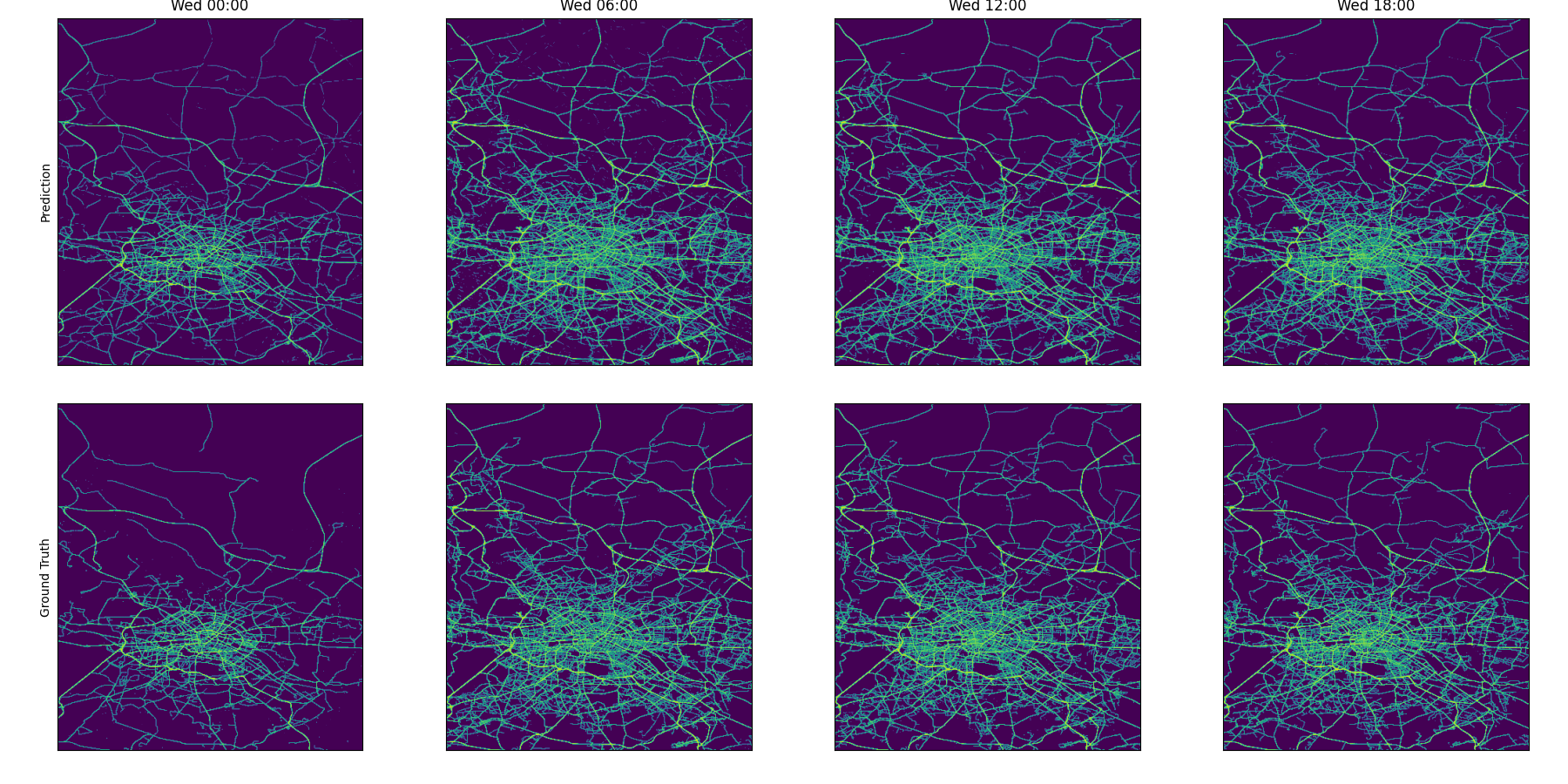}
    \caption{Prediction (top) and ground truth (bottom) for Berlin on Wed, 20.03.2019 at different times with logarithmic scale.}
    \label{fig:qualitative_results}
\end{figure*}

We evaluate our model using two different evaluation datasets. As the challenge test set is not openly available, we split the dataset and use a portion of the original training data only during model evaluation. The first evaluation set is a fraction of the training set ($\approx 0.4\%$) selecting specific points in time. Specifically, for each day in April 2019 (30 days), we sample the seed data at every full hour between 00:00 and 22:00 for each of the eight given cities.
Additionally, for exactly the same samples, we flip the data vertically and horizontally. This means that we flip the static images, as well as the dynamic traffic movies, and also rearrange the data channels accordingly. This results in a second dataset (denoted as mirrored) that is used to evaluate spatial generalization to new cities. The evaluation metric corresponds to the MSE between prediction and target image that are both scaled by 255 to match the original scale of the given data.

Fig. \ref{fig:qualitative_results} shows predictions and ground truth images for Berlin. The color values are log-scaled.

\begin{figure*}
    \centering
    \includegraphics[width=0.9\textwidth]{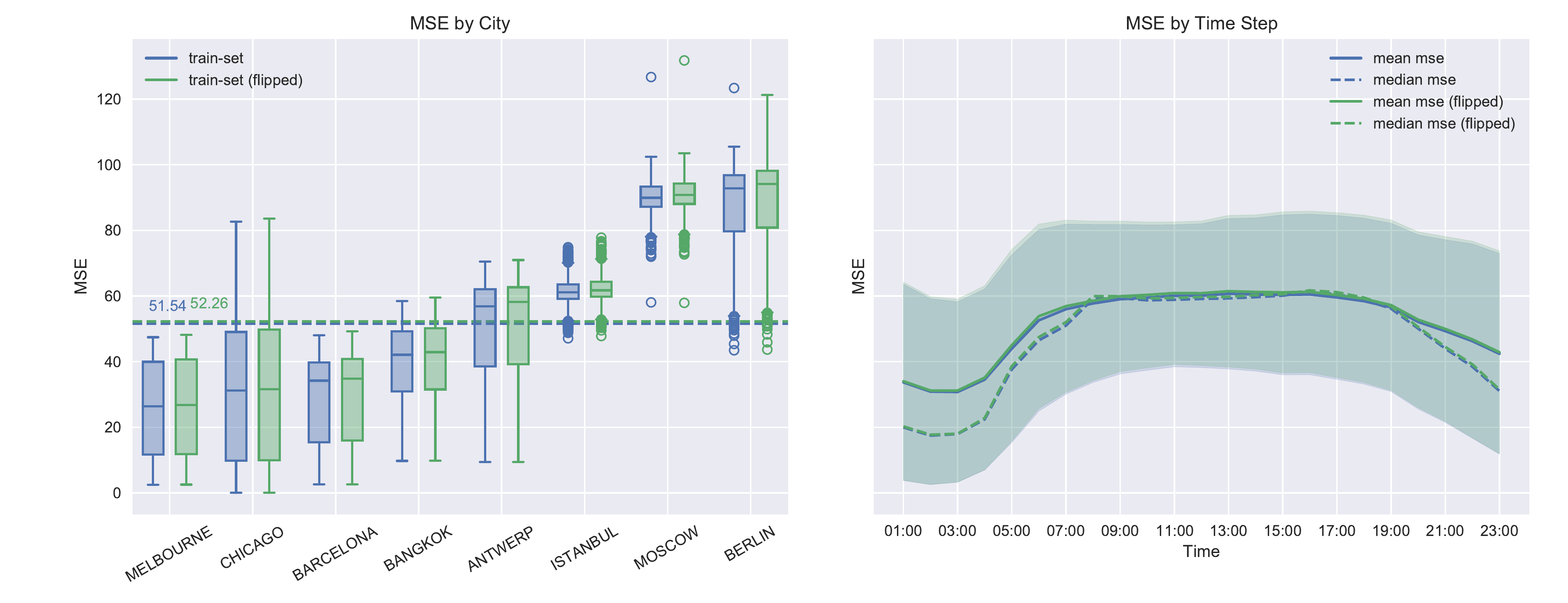}
    \caption{MSE by city (left) and by time (right) on both evaluation sets. Left: Dashed lines denote total MSE over the full test sets in the left plot. Right: Shaded areas denote the standard deviation; solid lines denote average; dashed lines denote median.}
    \label{fig:box_by_city}
\end{figure*}

Fig. \ref{fig:box_by_city} shows the MSE by city (left) and the MSE by time (right) for both evaluation sets. The average performance is also included (left plot; dashed lines). A high variation of performance across cities is observable. Furthermore, our model consistently performs better than average on the cities Melbourne and Barcelona, but consistently worse on the cities Istanbul, Moscow, and Berlin. 

The right plot in Fig. \ref{fig:box_by_city} shows that samples drawn at different points in time of a day cause variation in performance. Overall, predictions of samples drawn at nighttime, between 01:00 and 04:00 show a smaller error compared to predictions of samples drawn during daytime. This is probably due to the overall traffic activity, which is lower at nighttime.
The difference in performance measured on the two evaluation sets is very small, which demonstrates nicely the ability of the model to generalize to novel spatial situations.

Table \ref{tab:leaderboard} shows an excerpt of the leaderboard of the competition. The naive average corresponds to a model that calculates the average over the input frames and outputs the result for all future frames. Graph-ResNet by \cite{pmlr-v123-martin20a} was used as a more sophisticated baseline by the competition hosts. The graph-creation method was changed from using the dynamic data to build the graph, to using high-resolution street maps. The Graph-ResNet was trained for a single epoch only on traffic data from Berlin.
We ranked seventh place in the core- and fourth place in the extended challenge. In both, our model outperformed the baselines significantly. A comparison of performance between the two challenges is difficult, as different cities are used. However, we can compare the score ratio of our model to the baselines (s. relative score in Tab. \ref{tab:leaderboard}). the Graph-ResNet has a relative score of $0.841$, whereas our proposed model has $0.839$ relative score, which suggests better spatial generalization.


\begin{table}[]
    \centering
    \caption{Results taken from the competition leaderboard. The numbers correspond to the MSE, i.e. lower is better. Graph-ResNet BER was trained only on data from Berlin. T denotes temporal (core) challenge, ST denotes spatiotemporal (extended) challenge. Cited U-Net models differ between core (T) and extended competition (ST).
    }
    \label{tab:leaderboard}
    \resizebox{\linewidth}{!}{
    \begin{tabular}{l ll ll}
    \hline
     Model                 & \multicolumn{2}{c}{Competition} & & Rel. Score \\ 
                           & T    & ST & & T / ST \\ \toprule
    Naive Average
                           & 53.406       & 63.14        & & 0.846 \\ 
    Graph-ResNet BER \cite{pmlr-v123-martin20a}  
                           & 51.714       & 61.461       & & 0.841 \\ 
    Vanilla U-Net     \cite{ronnebergerUNetConvolutionalNetworks2015}
                           & 51.283       & -            & & - \\
    Hybrid U-Net (ours)    & 50.521       & 60.222       & & 0.839 \\ \hdashline
    U-Net + multi-task (T) \cite{u_net_lu} & \textbf{48.422}       & -            & & - \\
    U-Net + multi-task (ST) \cite{u_net_lu} & -            & 59.586       & & - \\
    U-Net Ensemble  (T)  \cite{t4c_choi_2019} & 48.494       & -       & & - \\
    U-Net Ensemble  (ST) \cite{t4c_choi_2019}  & -       & \textbf{59.559}       & & - \\
    \hline
    \end{tabular}
    }
\end{table}

\subsection{Ablation Study and Comparison to Vanilla U-Net}
To assess the effectiveness of the directional subgraphing, we train a model that is applied to the complete input graph instead of the four subgraphs. This is done simply by removing our modifications from the \textit{full GN Block}. Thereby, the graph operations become locally permutation invariant, instead of being sensitive to the neighborhood topology. We refer to this simplified version of our model as Graph U-Net.

Table \ref{tab:quantitative_results} shows the MSE measured on the evaluation dataset consisting of cities from the training set and on the mirrored evaluation dataset (denoted as MSE*). It can be observed that our Hybrid U-Net consistently outperforms the Graph U-Net on the evaluation dataset, whereas the results measured on the mirrored data are very similar for the two models.
To quantify spatial generalization, we compute the ratio between MSE and MSE* (denoted as rel. MSE). A relative MSE of $1.0$ means that the model produces exactly the same error on the mirrored cities as on the known cities. This would indicate that the performance is not impacted by the flipping of the data which corresponds to perfect spatial generalization.
The average rel. MSE measured for the Graph U-Net is consistently larger compared to the rel. MSE of our proposed Hybrid U-Net model. This indicates that Graph U-Net generalizes better to new cities than our Hybrid U-Net model.  



Table \ref{tab:quantitative_results} also shows the results for a vanilla U-Net. The vanilla U-Net consists of 8 consecutive down- and upsampling blocks, respectively. The MSE measured on the evaluation set is very close to our proposed model, but on average performs slightly worse. In contrast, the MSE* measured on the flipped cities is significantly worse, which is evidence of worse spatial generalization capabilities. This is further supported by the considerably worse relative MSE found for the vanilla U-Net.

\begin{table*}[]
    \centering
    \caption{MSE by city for the proposed Hybrid U-Net model, the Graph U-Net without subgraphing, a vanilla U-Net model. MSE* is measured on the mirrored dataset and rel. MSE is the ratio of MSE and MSE*. A rel. MSE of $1.0$ is reached if MSE = MSE*, which is an indicator for good spatial generalization. Bold numbers denote best performance on the respective evaluation dataset.}
    \label{tab:quantitative_results}
    \resizebox{\linewidth}{!}{
    \begin{tabular}{l lll l lll l lll}
    \hline
       & \multicolumn{3}{c}{Hybrid UNet} & & \multicolumn{3}{c}{Graph UNet} & & \multicolumn{3}{c}{Vanilla UNet} \\ 
    \cline{2-4}\cline{6-8}\cline{10-12}
            & MSE & MSE* & rel. MSE & & MSE & MSE* & rel. MSE & & MSE & MSE* & rel. MSE \\ \toprule
ANTWERP  	& 48.35 & \textbf{49.034} & 0.986 & & 48.819 & 49.186 & 0.993 & & \textbf{48.193} & 50.712 & 0.95 \\ 
BANGKOK  	& 39.466 & 40.338 & 0.978 & & 39.729 & \textbf{40.045} & 0.992 & & \textbf{39.444} & 40.908 & 0.964 \\ 
BARCELONA  	& 28.742 & 29.502 & 0.974 & & 28.968 & \textbf{29.284} & 0.989 & & \textbf{28.609} & 29.663 & 0.964 \\ 
BERLIN  	& 87.047 & 88.41 & 0.985 & & 87.798 & \textbf{88.388} & 0.993 & & \textbf{86.95} & 91.068 & 0.955 \\ 
CHICAGO  	& \textbf{32.147} & 32.593 & 0.986 & & 32.451 & \textbf{32.526} & 0.998 & & 32.228 & 32.939 & 0.978 \\ 
ISTANBUL  	& \textbf{61.237} & \textbf{62.028} & 0.987 & & 61.98 & 62.262 & 0.995 & & 61.588 & 64.3 & 0.958 \\ 
MELBOURNE  	& \textbf{25.325} & 25.74 & 0.984 & & 25.626 & \textbf{25.709} & 0.997 & & 25.393 & 26.091 & 0.973 \\ 
MOSCOW  	& \textbf{89.628} & \textbf{90.587} & 0.989 & & 90.44 & 90.855 & 0.995 & & 89.846 & 93.752 & 0.958 \\ 
     	 \hdashline
\textit{average}  	& \textbf{51.493} & \textbf{52.279} & 0.985 & & 51.976 & 52.282 & 0.994 & & 51.531 & 53.679 & 0.96 \\ 
    \hline
    \end{tabular}
    }
\end{table*}


\section{Discussion and Conclusion}
The given problems in the \textit{Traffic4Cast} challenge have been approached following two different routes. Either using a visual model to process whole frames of the traffic movies or using a GNN and process only pixels that actually depict a road.
Intuitively, a graph-based approach is leveraging prior knowledge of the underlying structure of the street network, which should provide better generalization and transfer. Additionally, areas without streets are excluded from the graph and therefore don't explicitly contribute to the prediction. This is intuitively beneficial as these areas don't contain traffic information. 
This has, in principle, already been demonstrated by \cite{pmlr-v123-martin20a}. But as a drawback, a purely graph-based approach is losing information on directionality which is crucial for the traffic forecasting challenge as the data is provided in such a format.

Here, we introduced a U-Net architecture with graph layers that we adapted to be sensitive to the geographical neighborhood topology by splitting the road graph into four direction-dependent subgraphs. Furthermore, we utilize the 2D node position for graph down- and upsampling which effectively expands the receptive field and allows inference based on a larger portion of the road network.
Our model, much like the one by \cite{pmlr-v123-martin20a}, showed improved spatial generalization compared to a convolutional model. Furthermore, our results indicate that biasing the graph operation with geographical topology improves the overall performance while not sacrificing much of the spatial generalization capabilities.

Although we have shown that the approach works in general, we did not perform extensive hyperparameter tuning, yet. Investigating the difference in performance between the different cities, refining the proposed up- and downsampling layers, and scaling the complexity of the model 
are promising directions to explore in future work.

\bibliographystyle{unsrtnat}
\bibliography{references}  






\end{document}